\documentclass[journal]{IEEEtran}

\usepackage{xcolor}
\usepackage{amssymb}
\usepackage{siunitx}

\usepackage{cite}
\ifCLASSINFOpdf
   \usepackage[pdftex]{graphicx}
   \graphicspath{{/Figures/}}
\else
\fi
\usepackage{amsmath}
\usepackage{bm}
\begin{document}
\title{Lensless Imaging with Compressive Ultrafast Sensing}
\author{Guy~Satat,~%~\IEEEmembership{Student~Member,~IEEE,}
		Matthew~Tancik
        and~Ramesh~Raskar%,~\IEEEmembership{Member,~IEEE,}% <-this % stops a space
\thanks{G. Satat, M. Tancik and R. Raskar are with the Media Lab, Massachusetts Institute of Technology, Cambridge, MA, 02139 USA. e-mail: guysatat@mit.edu.}% <-this % stops a space
}




\maketitle

%%%%%%%%%%%%%%%%%%%%%%%%%%%%%%%%%%%%%%%%%%%%%%%%%%%%%%%%%%%%%%%%%%%%%%%%%%%%%%%%%%%%%%%%%%%%%%%%%%%%%%%%%%%%%%%%%%%%%%%%%%%%%%%%%%%%%%%%%%%%%%%%%%%%%%%%%%%%%%%%%%%%%%%%%%%%%%%%%%%%%%%%%%%%%%
\begin{abstract}
Lensless imaging is an important and challenging problem. One notable solution to lensless imaging is a single pixel camera which benefits from ideas central to compressive sampling. However, traditional single pixel cameras require many illumination patterns which result in a long acquisition process. Here we present a method for lensless imaging based on compressive ultrafast sensing. Each sensor acquisition is encoded with a different illumination pattern and produces a time series where time is a function of the photon's origin in the scene. Currently available hardware with picosecond time resolution enables time tagging photons as they arrive to an omnidirectional sensor. This allows lensless imaging with significantly fewer patterns compared to regular single pixel imaging. To that end, we develop a framework for designing lensless imaging systems that use ultrafast detectors. We provide an algorithm for ideal sensor placement and an algorithm for optimized active illumination patterns. We show that efficient lensless imaging is possible with ultrafast measurement and compressive sensing. This paves the way for novel imaging architectures and remote sensing in extreme situations where imaging with a lens is not possible.
\end{abstract}

%\begin{IEEEkeywords}
%IEEE, IEEEtran, journal, \LaTeX, paper, template.
%\end{IEEEkeywords}

% \fi
\IEEEpeerreviewmaketitle

%%%%%%%%%%%%%%%%%%%%%%%%%%%%%%%%%%%%%%%%%%%%%%%%%%%%%%%%%%%%%%%%%%%%%%%%%%%%%%%%%%%%%%%%%%%%%%%%%%%%%%%%%%%%%%%%%%%%%%%%%%%%%%%%%%%%%%%%%%%%%%%%%%%%%%%%%%%%%%%%%%%%%%%%%%%%%%%%%%%%%%%%%%%%%%
\section{Introduction}
\IEEEPARstart{T}{raditional} imaging is based on lenses that map the scene plane to the sensor plane. In this physics-based approach the imaging quality depends on parameters such as lens quality, numerical aperture, density of the sensor array and pixel size. Recently it has been challenged by modern signal processing techniques. Fundamentally the goal is to transfer most of the burden of imaging from high quality hardware to computation. This is known as computational imaging, in which the measurement encodes the target features; these are later computationally decoded to produce the desired image. Furthermore, the end goal is to completely eliminate the need for high quality lenses, which are heavy, bulky, and expensive.

One of the key workhorses in computational imaging is compressive sensing (CS)~\cite{Donoho2006a,Candes2006}. For example, CS enabled the single pixel camera \cite{Duarte2008a}, which demonstrated imaging with a single pixel that captured scene information encoded with a spatial light modulator (SLM). The pixel measurement is a set of consecutive readings, with different SLM patterns. The scene is then recovered using compressive deconvolution.

Broadly, the traditional imaging and single pixel camera demonstrate two extremes: traditional cameras use a pure hardware approach whereas single pixel cameras minimize the requirement for high quality hardware using modern signal processing. There are many trade-offs between the two approaches. One notable difference is the overall acquisition time: the physics-based approach is done in one shot (i.e. all the sensing is done in parallel). The single pixel camera and its variants require hundreds of consecutive acquisitions, which translates into a substantially longer overall acquisition time. 

Recently, time-resolved sensors enabled new imaging capabilities. Here we consider a time-resolved system with pulsed active illumination combined with a sensor with a time resolution on the order of picoseconds. Picosecond time resolution allows distinguishing between photons that arrive from different parts of the target with $mm$ resolution. The sensor provides more information per acquisition (compared to regular pixel), and so fewer masks are needed. Moreover, the time-resolved sensor is characterized by a measurement matrix that enables us to optimize the active illumination patterns and reduce the required number of masks even further. 

Currently available time-resolved sensors allow a wide range of potential implementations. For example, Streak cameras provide picosecond or even sub-picosecond time resolution~\cite{scheidt2000review}, however they suffer from poor sensitivity. Alternatively, Single Photon Avalanche Photodiode (SPAD) are compatible with standard CMOS technology~\cite{Richardson2009} and allow time tagging with resolutions on the order of tens of picoseconds. These devices are available as a single pixel or in pixel arrays. 

In this paper we present a method that leverages both time-resolved sensing and compressive sensing. The method enables lensless imaging for reflectance recovery with fewer illumination patterns compared to traditional single pixel cameras. This relaxed requirement translates to a shorter overall acquisition time. The presented framework provides guidelines and decision tools for designing time-resolved lensless imaging systems. In this framework the traditional single pixel camera is one extreme design point which minimizes the cost with simple hardware, but requires many illumination patterns (long acquisition time). Better hardware reduces the acquisition time with fewer illumination patterns at the cost of complexity. We provide sensitivity analysis of reconstruction quality to changes in various system parameters. 
Simulations with system parameters chosen based on currently available hardware indicate potential savings of up to  $50 \times$ fewer illumination patterns compared to traditional single pixel cameras.

\subsection{Contributions}
The contributions presented here can be summarized as:
\begin{enumerate}
\item Computational imaging framework for lensless imaging with compressive time-resolved measurement,
\item Analysis of a time-resolved sensor as an imaging pixel,
\item Algorithm for ideal sensor placement in a defined region,
\item Algorithm for optimized illumination patterns.
\end{enumerate}

\subsection{Related Works}

\subsubsection{Compressive Sensing for Imaging}
Compressive sensing has inspired many novel imaging modalities. Examples include: ultra-spectral imaging~\cite{August2016}, subwavelength imaging~\cite{Szameit2012}, wavefront sensing~\cite{Polans2014}, holography~\cite{Brady2009}, imaging through scattering media~\cite{Liutkus2014a}, terahertz imaging~\cite{Watts2014}, and ultrafast imaging~\cite{bosworth2015high,gao2014single}.

\subsubsection{Single Pixel Imaging}
One of the most notable applications of compressive sensing in imaging is the single pixel camera~\cite{Duarte2008a}. This was later extended to general imaging with masks~\cite{Bahmani2015}. We refer the interested reader to an  introduction on imaging with compressive sampling in~\cite{Romberg2008}. 

Other communities have also discussed the use of indirect measurements for imaging. In the physics community the concept of using a single pixel (bucket) detector to perform imaging is known as ghost imaging and was initially thought of as a quantum phenomenon~\cite{Pittman1995}. It was later realized that computational techniques can achieve similar results~\cite{Shapiro2008a}. Ghost imaging was  also incorporated with compressive sensing~\cite{Katz2009,Katkovnik2012}. In the computational imaging community this is known as dual photography~\cite{Sen2005}.

Single pixel imaging extends to multiple sensors. For example, multiple sensors were used for 3D reconstruction of a scene by using stereo reconstruction~\cite{Sun2013}. Multiple sensors were also incorporated with optical filters to create color images~\cite{Welsh2013}.  

In this work we suggest using a time-resolved sensor instead of a regular bucket detector for lensless imaging.

\subsubsection{Time-Resolved Sensing for Imaging}
Time-resolved sensing has been mostly used to recover scene geometry. This is known as LIDAR~\cite{Schwarz2010}. LIDAR was demonstrated with a compressive single pixel approach~\cite{Kirmani2011, colacco2012compressive}.
Time-resolved sensing has also been suggested to recover scene reflectance~\cite{Wu2014,Kirmani2012} for lensless imaging, but without the use of structured illumination and compressive sensing. 
Other examples of time-resolved sensing include non-line of sight imaging, for example imaging around a corner~\cite{Velten2012} and through scattering~\cite{Satat2015,Satat2016all}. Imaging around corners was also demonstrated with low cost time-of-flight sensors, using back propagation~\cite{kadambi2016occluded} and sparsity priors~\cite{heide2014diffuse}.

In this paper we use compressive deconvolution with time-resolved sensing for lensless imaging to recover target reflectance.

\subsection{Limitations}
The main limitations of using our suggested approach are:
\begin{itemize}
\item We assume a linear imaging model (linear modeling in imaging is common, for example \cite{Duarte2008a, Bahmani2015}).
\item Our current implementation assumes a planar scene. We note that our approach can naturally extend to piecewise planar scenes and leave this extension to a future study.
\item Time-resolved sensing requires an active pulsed illumination source and a time-resolved sensor. These can be expensive and complicated to set up. However, as we demonstrate here, they provide a different set of trade-offs for lensless imaging, primarily reduced acquisition time.
\end{itemize}

%%%%%%%%%%%%%%%%%%%%%%%%%%%%%%%%%%%%%%%%%%%%%%%%%%%%%%%%%%%%%%%%%%%%%%%%%%%%%%%%%%%%%%%%%%%%%%%%%%%%%%%%%%%%%%%%%%%%%%%%%%%%%%%%%%%%%%%%%%%%%%%%%%%%%%%%%%%%%%%%%%%%%%%%%%%%%%%%%%%%%%%%%%%%%%
\section{Compressive Ultrafast Sensing}
\label{sec:probelm_setup}

\begin{figure}[t]
\centering
\includegraphics[width=\linewidth]{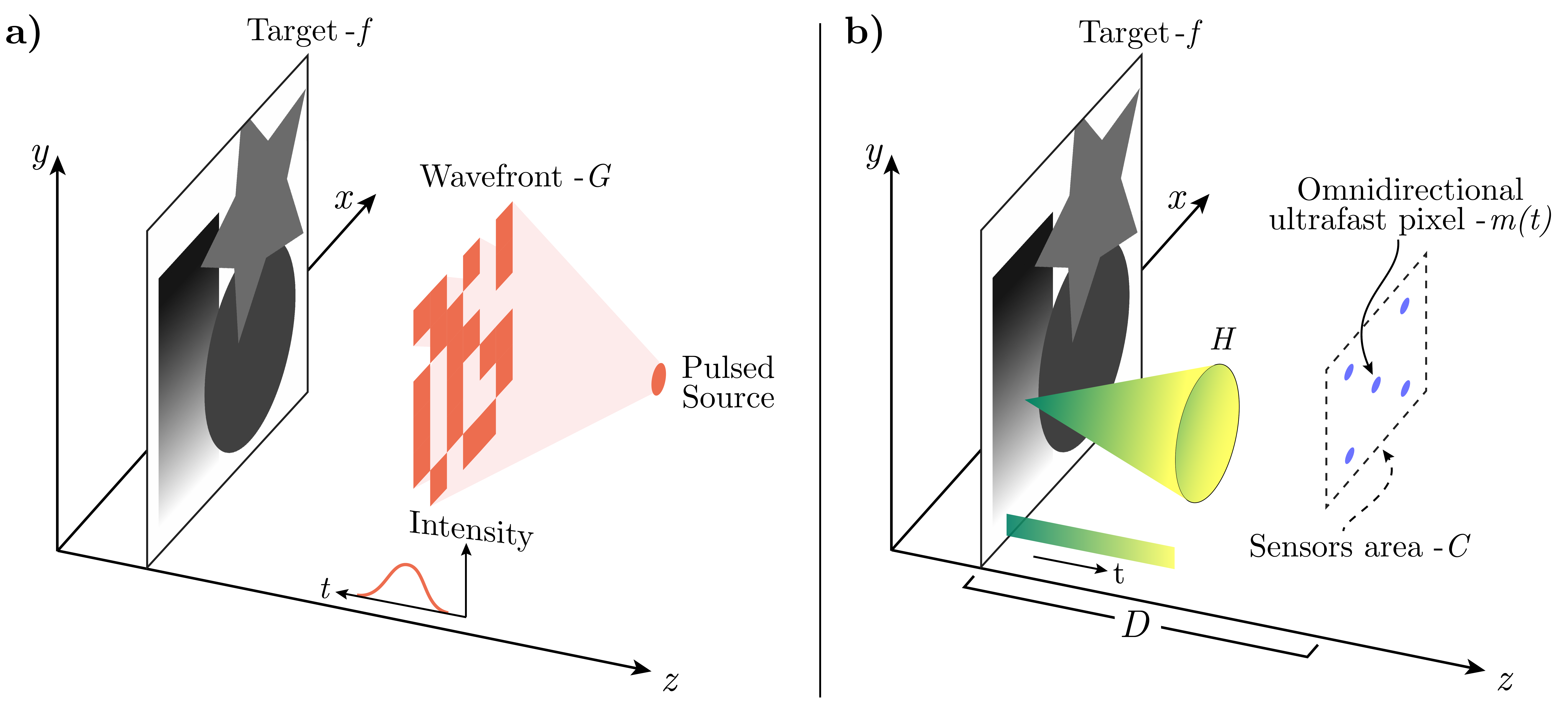}
\caption{Lensless imaging with compressive ultrafast sensing. a) Illumination, a time pulsed source is wavefront modulated ($\mathbf{G}$) and illuminates a target with reflectance $f$. b) Measurement, omnidirectional ultrafast sensor (or sensors) measures the time dependent response of the scene $m(t)$. $\mathbf{H}$ is a physics-based operator that maps scene pixels to the time-resolved measurement.}
\label{fig:setup_view}
\end{figure}

Our goal is to develop a framework for compressive imaging with time-resolved sensing. Fig.~\ref{fig:setup_view} shows the system overview. A target $\mathbf{f} \in \mathbb{R}^L$ with $L$ pixels is illuminated by wavefront $\mathbf{G} \in \mathbb{R}^L$. $\mathbf{G}$ is produced by spatially modulating a time pulsed source. Light reflected from the scene is measured by an omnidirectional ultrafast sensor with time resolution $T$ positioned on the sensors plane. The time-resolved measurement is denoted by $\mathbf{m} \in \mathbb{R}^N$, where $N$ is the number of time bins in the measurement. Better time resolution (smaller $T$) increases~$N$. $\mathbf{H} \in \mathbb{R}^{N \times L}$ is the measurement matrix defined by the space to time mapping that is enforced by special relativity. In the case when the time resolution is very poor, $\mathbf{H}$ is just a single row ($N=1$), and the process is reduced to the regular single pixel camera case.

We consider $K$ sensors ($i=1..K$) with $N$ time samples and $M$ illumination patterns ($j=1..M$) so the time-resolved measurement of the $i$-th sensor, for a target illuminated by the $j$-th illumination pattern, is defined by: ${\mathbf{m}_{i,j}=\mathbf{H}_i\mathbf{G}_j\mathbf{f} }$. Concatenating all measurement vectors results in the total measurement vector $\vec {\mathbf{m}} \in \mathbb{R}^{NKM}$, such that the total measurement process is:
\begin{equation}
\vec {\mathbf{m}} = \left[ {\begin{array}{*{20}{c}}
 \vdots \\
{{\mathbf{m}_{i,j}}}\\
 \vdots 
\end{array}} \right] = \left[ {\begin{array}{*{20}{c}}
 \vdots \\
{{\mathbf{H}_i}{\mathbf{G}_j}}\\
 \vdots 
\end{array}} \right]\mathbf{f} = \mathbf{Q}\mathbf{f}
\label{eq:mesurement_operator}
\end{equation}
where, $\mathbf{Q}$ is an $NKM \times L$ matrix which defines the total measurement operator. 

Here we invert the system defined in Eq.~\ref{eq:mesurement_operator} using compressive sensing approach. 
To that end, we analyze and physically modify $\mathbf{Q}$ to make the inversion robust. In the remainder of this paper we analyze and optimize the following fundamental components of $\mathbf{Q}$:
\begin{itemize}
\item Physics-based time-resolved light transport matrix $\mathbf{H}$. $\mathbf{H}$~is a mapping from the spatial coordinates of the  scene domain to the time-resolved measurement ($\mathbf{H}: \mathbf{r} \to t$). Section \ref{sec:time_resolved_light_transport} derives a physical model of $\mathbf{H}$ and discusses its structure and properties. $\mathbf{H}$ can be modified by changing the sensor time resolution and position in the sensor's plane.

\item Combination of multiple sensors. Multiple sensors can be placed in the sensor plane, such that each sensor will correspond to a different time-resolved light transport matrix $\mathbf{H}_i$. Section \ref{sec:optimized_array} presents an algorithm for optimized sensor placement in the sensor's plane.
 
\item Illumination (probing) matrix $\mathbf{G}$. This matrix is similar to the sensing matrix in the single pixel camera case which was realized then with an SLM. In our analysis we assume the modulation is performed on the illumination side (but note that modulating the illumination is equivalent to modulating the incoming wavefront to the sensor). The structured illumination allows modulating the illumination amplitude on different pixels in the target. 
%We consider multiple illumination patterns such that $\mathbf{G}_j$ corresponds to the $j$-th illumination pattern. 
Section \ref{sec:optimized_illum} presents an algorithm for optimized illumination patterns for compressive ultrafast imaging.
\end{itemize} 

The inversion of Eq.~\ref{eq:mesurement_operator} is robust if there is little linear dependence among the columns of $\mathbf{Q}$ (so that it has sufficient numerical rank). This is evaluated by the mutual coherence~\cite{Elad2007} which is a measure for the worst similarity of the matrix columns and is defined by:
\begin{equation}
\mu  = \mathop {\max }\limits_{1 \le a,b \le {L},a \ne b} \frac{{\left| {{\mathbf{Q}_a}^T{\mathbf{Q}_b}} \right|}}{{\left\| {{\mathbf{Q}_a}} \right\|_2  \left\| {{\mathbf{Q}_b}} \right\|_2}}
\label{eq:coherence}
\end{equation}
From here on, as suggested in \cite{Duarte-Carvajalino2009}, we will use an alternative way to target the mutual coherence which is computationally tractable and defined by:
\begin{equation}
 \mu  = \frac{1}{{{L}}}{\left\| {{\mathbf{I}_{{L}}} - {{ \mathbf{\tilde Q}}^T} \mathbf{\tilde Q}} \right\|_F^2}
\label{eq:cohrence_measure}
\end{equation}
where $\mathbf{I}_{{L}}$ is the identity matrix of size $L$, $ \mathbf{\tilde Q}$ is $\mathbf{Q}$ with columns normalized to unity, and ${\left\|  \cdot  \right\|_F}$ is the Frobenius norm. This definition directly targets the restricted isometry property (RIP)~\cite{Duarte-Carvajalino2009}, which provides guarantees for using compressive sensing. In the remainder of the paper we optimize different parts of $\mathbf{Q}$ using this measure of coherence as a cost objective.

\begin{figure}
\centering
\includegraphics[width=.95\linewidth]{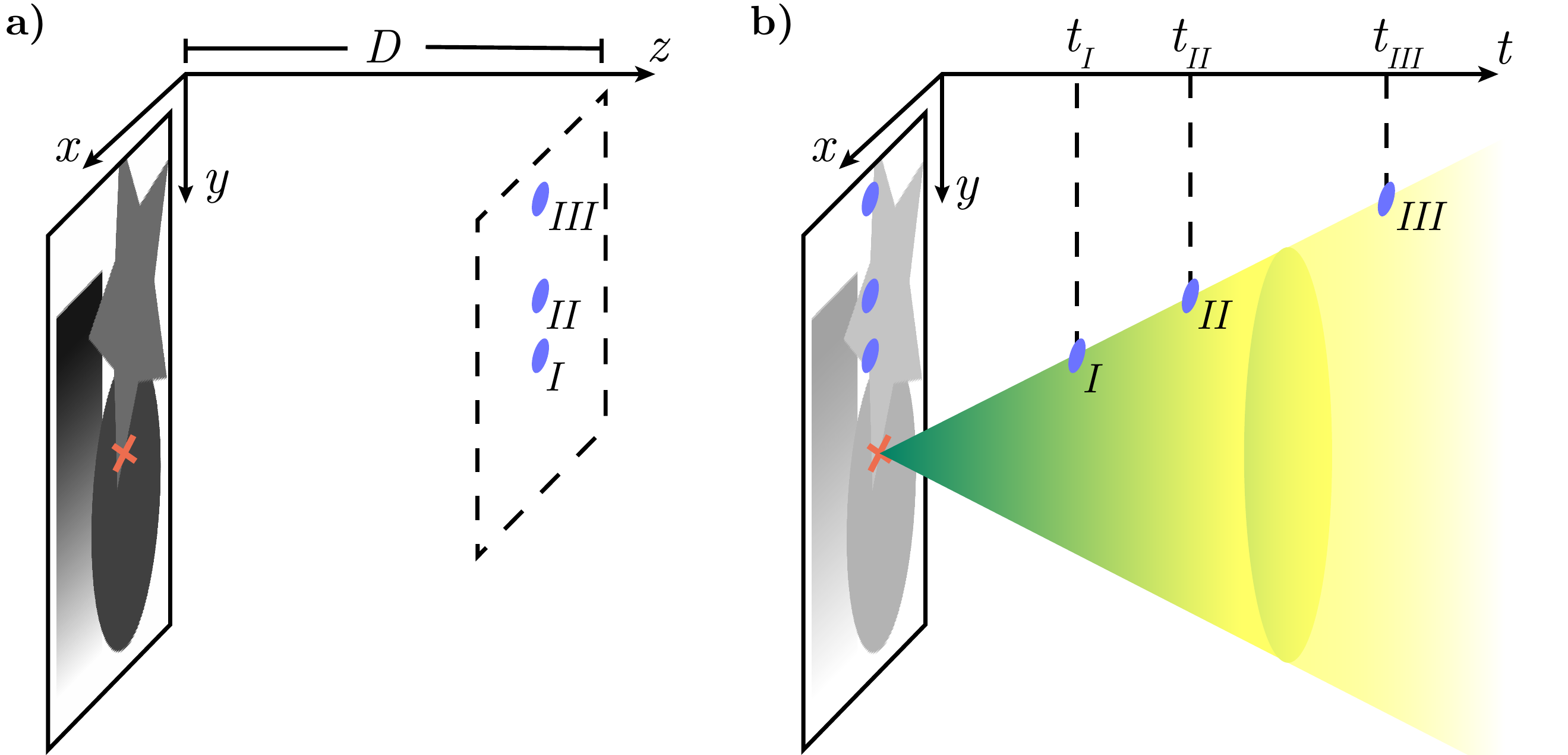}
\caption{Schematic of the light cone for the case of a planar stationary target. a) Scene geometry, the target plane and sensor plane are separated by a distance $z=D$. Three detectors are positioned at different $y$ positions in the detector plane. b) The light-like part of the light cone emanating from the target point marked with a red 'X' defines the measurement times of the different detectors. Due to the geometry, the light cone will arrive to the detectors at different times. First it will be measured by detector I which is closest to the source, followed by detector II and III.   }
\label{fig:light_cone}
\end{figure}

%%%%%%%%%%%%%%%%%%%%%%%%%%%%%%%%%%%%%%%%%%%%%%%%%%%%%%%%%%%%%%%%%%%%%%%%%%%%%%%%%%%%%%%%%%%%%%%%%%%%%%%%%%%%%%%%%%%%%%%%%%%%%%%%%%%%%%%%%%%%%%%%%%%%%%%%%%%%%%%%%%%%%%%%%%%%%%%%%%%%%%%%%%%%%%
\section{Time-Resolved Light Transport}
\label{sec:time_resolved_light_transport}

We start by developing a generic light transport model for time-resolved imaging. The finite speed of light governs information propagation, and provides geometrical constraints which will be used in the image formation model. These are conveniently described in a Minkowski space with the space-time four-vector $(\mathbf{r},t)=(x,y,z,t)$. If we consider a point source at position $\mathbf{r}'$ pulsing at time $t'$, and a sensor at position $\mathbf{r}$, then the space-time interval between the source $(\mathbf{r}',t')$ and the sensor $(\mathbf{r},t)$ is defined by:
\begin{equation}
{s^2} = {\left\| {\mathbf{r} - \mathbf{r}'} \right\|_2^2} - {c^2}{(t - t')^2}
\label{eq:four_vector_norm}
\end{equation}
where $c$ is the speed of light. Enforcing causality and light-like behavior requires $s^2=0$ which defines the light cone. Fig.~\ref{fig:light_cone} shows a schematic of the light cone and demonstrates how the same event is measured in various positions at different times. 

Thus, the time-resolved measurement of a sensor at position $\mathbf{r}$ and time $t$ of a general three-dimensional time dependent scene $f(\mathbf{r}',t')$ is the integral over all $(\mathbf{r}',t')$ points on the manifold $M$ defined by ${s^2={\left\| {\mathbf{r} - \mathbf{r}'} \right\|_2^2} - {c^2}{(t - t')^2}=0}$:
\begin{equation}
m\left( {{\bf{r}},t} \right) = \int\limits_{M}^{} {\frac{1}{{\left\| {{\bf{r}} - {\bf{r'}}} \right\|_2^2}}f({\bf{r'}},t')dM} 
\label{eq:general_measurement}
\end{equation}
where the ${1/{\left\| \mathbf{r}-\mathbf{r}' \right\|_2^2}}$ term accounts for intensity drop off.

\begin{figure}[t]
\centering
\includegraphics[width=\linewidth]{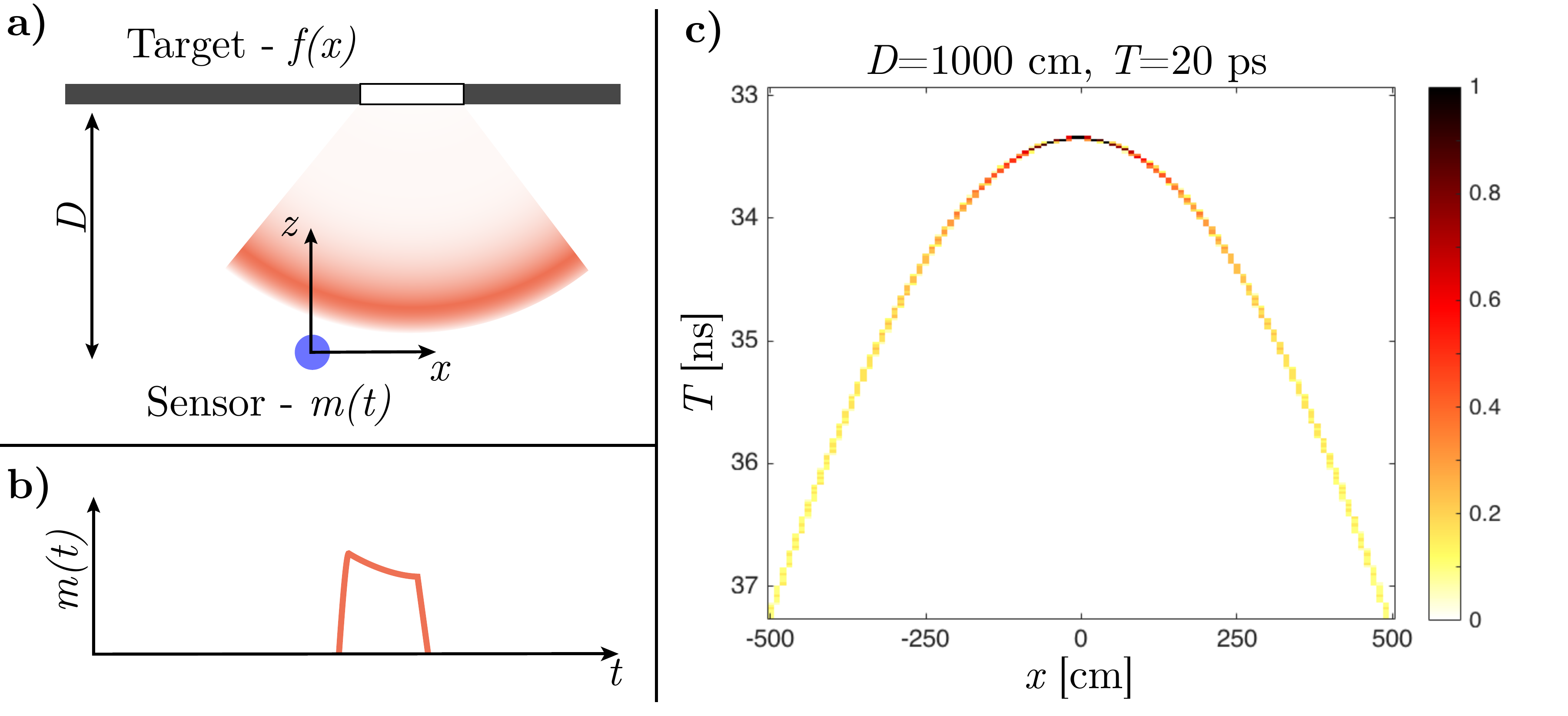}
\caption{Analysis of a one-dimensional world. a) Geometry, the target is a black line with a white patch, at a distance $D$ from the time-resolved sensor. b)~The time-resolved measurement produced by the sensor. The signal start time corresponds to the patch distance, and the time duration to the patch width. c)~The measurement matrix $\mathbf{H}$, generated from Eq. \ref{eq:one_d_measurement}. Here the distance to the target is $D=\SI{1000}{\centi\meter}$ and the sensor has a time resolution of ${T=\SI{20}{\pico\second}}$.}
\label{fig:one_d_world}
\end{figure}

Next, we assume a planar scene at $z'=D$, sensor at $z=0$ and a stationary target so that $t'$ is fixed and can be assumed ${t'=0}$ without loss of generality. $\mathbf{f}$ is a discretized, lexicographically ordered representation of the target reflectance map $f(x,y)$. We use the circular symmetry of the light cone so that:
\begin{equation}
m(x,y,t) = \int\limits_0^{2\pi } {\frac{1}{{{c^2}{t^2}}}f\left( {x + \rho \cos (\theta '),y + \rho \sin (\theta ')} \right)\rho d\theta '} 
\label{eq:general_measurement_with_angle}
\end{equation}
with ${\rho  = \sqrt {{c^2}{t^2} - {D^2}}} $. The intensity drop off is written as a function of time due to: ${{\left\| {\mathbf{r} - \mathbf{r}'} \right\|}_2^2}=c^2 \left(t-t' \right)^2=c^2 t^2$. Fig.~\ref{fig:light_cone} shows a schematic of the light cone for this case.

The sensor's finite time resolution $T$ corresponds to the sampling of $m(x,y,t)$ and denoted by $\mathbf{m}$. A sensor positioned at location $\mathbf{r}_i=(x_i,y_i)$ will produce a measurement ${\mathbf{m}_i=\mathbf{H}_i \mathbf{f}}$, where $\mathbf{H}_i$ is defined by the kernel in Eq. \ref{eq:general_measurement_with_angle}. $\mathbf{H}_i$ is a mapping from a two-dimensional spatial space to a time measurement (dependent on the detector position). The kernel maps rings with varying thicknesses from the scene plane to specific time bins in the measurement. The next subsection discusses the properties of this kernel.

\subsection{One-Dimensional Analysis}

It is interesting to analyze $\mathbf{H}$ in a planar world ($y=0$) with the sensor at the origin (Fig. \ref{fig:one_d_world}). In that case Eq. \ref{eq:general_measurement_with_angle} is simplified to:
%\begin{equation}
%m(t) = \frac{1}{{{c^2}{t^2}}}\left[ {f\left( {\sqrt {{c^2}{t^2} - {D^2}} } \right)} \right]
%\label{eq:one_d_measurement}
%\end{equation}
\begin{equation}
m(t) = \frac{1}{{{c^2}{t^2}}}\left[ {f\left( { - \sqrt {{c^2}{t^2} - {D^2}} } \right) + f\left( {\sqrt {{c^2}{t^2} - {D^2}} } \right)} \right]
\label{eq:one_d_measurement}
\end{equation}
Fig.~\ref{fig:one_d_world}c shows an example of the corresponding $\mathbf{H}$ matrix. This simple example demonstrates the key properties of the time-resolved measurement: 1) It is a nonlinear mapping of space to time. 2) The mapping is not unique (two opposite space points are mapped to the same time slot). 3) Spatial points that are close to the sensor are undersampled (adjacent pixels mapped to the same time slot). 4) Spatial points that are far from the sensor are oversampled but they are of a weaker signal. These properties affect imaging parameters as described next:

\subsubsection{Resolution limit} 
\begin{figure}[t]
\centering
\includegraphics[width=\linewidth]{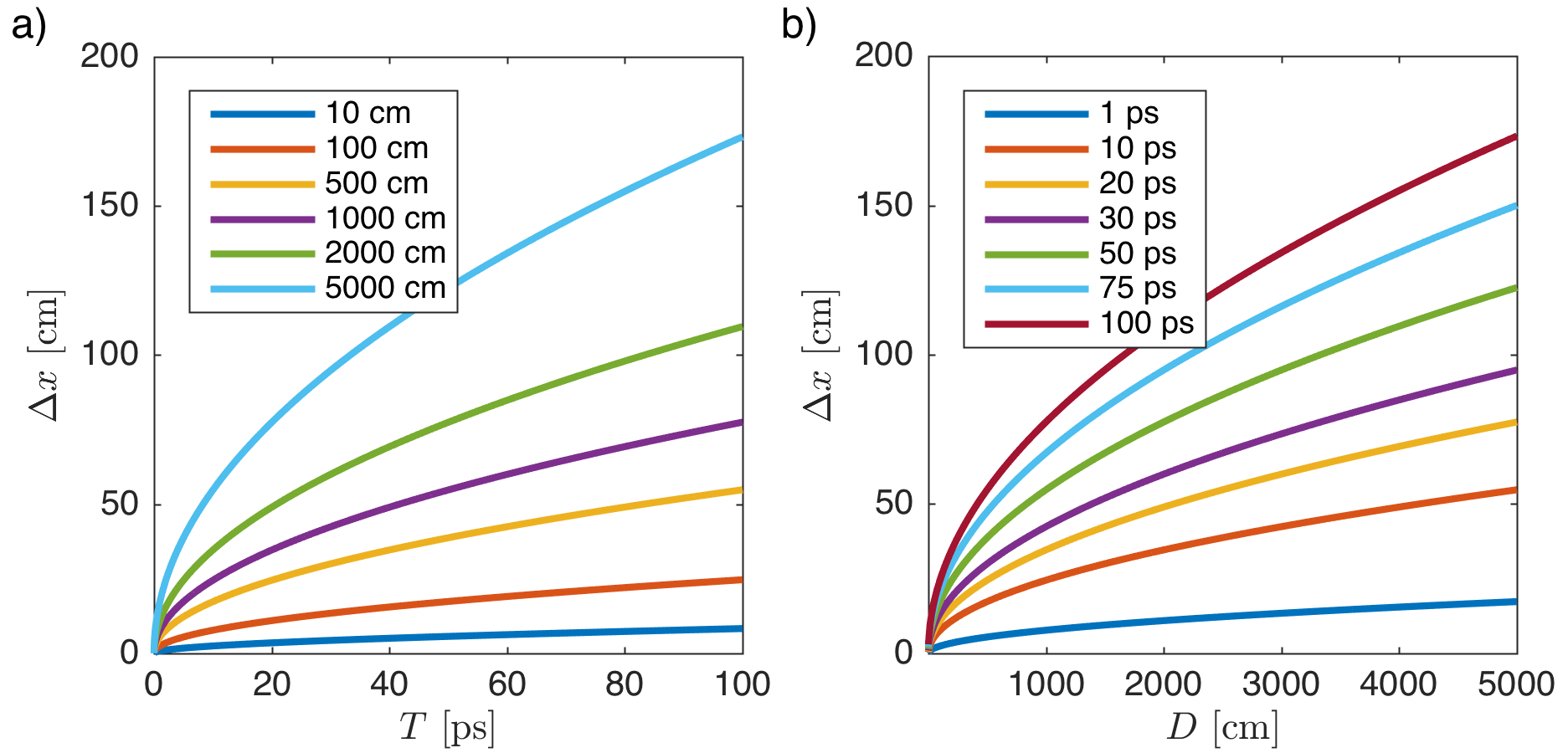}
\caption{Recoverable resolution with time-resolved sensing. a) Plots for various scene distances $D$ as a function of sensor time resolution $T$. b) Plots for various sensor time resolutions as a function of target distance $D$.}
\label{fig:one_d_world_res}
\end{figure}

The minimum recoverable spatial resolution is defined by the closest point to the sensor: ${ct_1=D}$, and the point that corresponds to the next time slot: ${c(t_1+T)=\sqrt{D^2+\Delta x^2}}$, which results in :
\begin{equation}
\Delta x = cT\sqrt{1+2\frac{D}{cT}}
\label{eq:one_d_min_res}
\end{equation}  
Fig. \ref{fig:one_d_world_res} shows a few cross sections of Eq. \ref{eq:one_d_min_res} for relevant distances $D$ and time resolution $T$. Better time resolution is required for further scenes (for the same recoverable resolution).

\subsubsection{Signal to noise ratio and dynamic range limitation} The closest point to the sensor defines the measurement gain (in order to avoid the saturation intensity $I_{sat}$), such that ${I_{sat}>A \left( {D^2} \right)^{-1}}$, where $A$ accounts for all measurement constants. The furthest measurable point from the sensor ($x_{max}$) should result in a measurement above the noise floor: $I_{n}<A\left( {D^2+x_{max}^2} \right)^{-1}$. 

Since the signal to noise ratio (SNR) is proportional to the sensor gain, closer scenes will have smaller coverage areas. For example, if we assume: ${I_{sat}=B I_n}$ (for some constant $B>1$) we get: ${x_{max}<\sqrt{B-1}}~{D}$.

\begin{figure}[t]
\centering
\includegraphics[width=\linewidth]{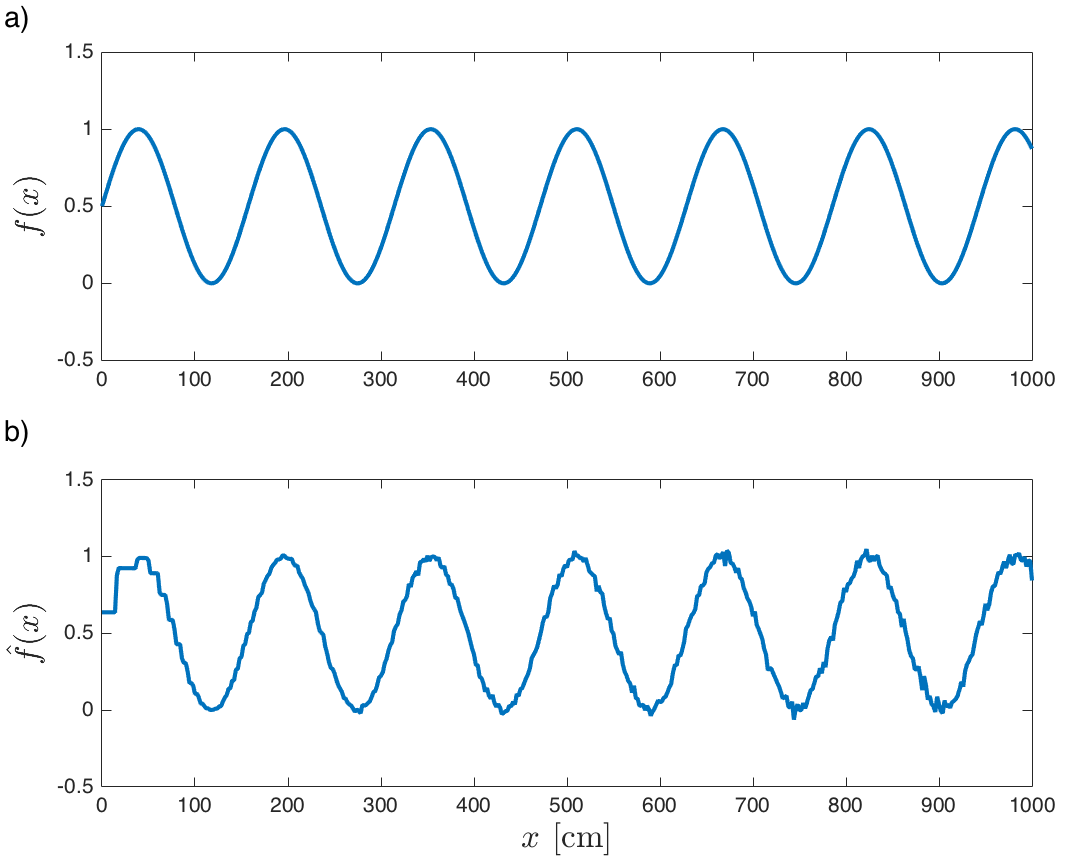}
\caption{Effects of averaging and noise on time-resolved measurement. a)~$f(x)$ is a sinusoid on the positive half plane, at a distance $D=\SI{1000}{\centi\meter}$ from a sensor with time resolution $T=\SI{20}{\pico\second}$ and measurement noise of SNR${=\SI{35}{\deci\bel}}$. b) $\hat{f}(x)$ is the result of inverting the system using the Moore-Penrose pseudoinverse, which demonstrates the undersampled measurement close to the sensor, and sensitivity to noise further away from the sensor.}
\label{fig:one_d_world_sin}
\end{figure}

The combined effect of these phenomena is demonstrated in Fig. \ref{fig:one_d_world_sin}. In this example, we consider a `half plane' ($x>0$), where the target reflectance is ${f(x)=sin(x)}$ with additive white Gaussian noise, which results in  SNR${=\SI{35}{\deci\bel}}$. For this simple demonstration, we use the Moore-Penrose pseudoinverse to invert the system, such that ${\mathbf{\hat f} = \mathbf{H}^\dagger \mathbf{m}}$. The inversion shows that close to the origin ($x<\SI{50}{\centi\meter}$) the reconstruction suffers from an undersampled measurement; this area is not sensitive to the measurement noise, and looks identical with zero noise. The noise has an obvious effect on the reconstruction further from the origin ($x>\SI{700}{\centi\meter}$).

\subsection{Analysis of a planar scene}
\begin{figure}[t]
\centering
\includegraphics[width=\linewidth]{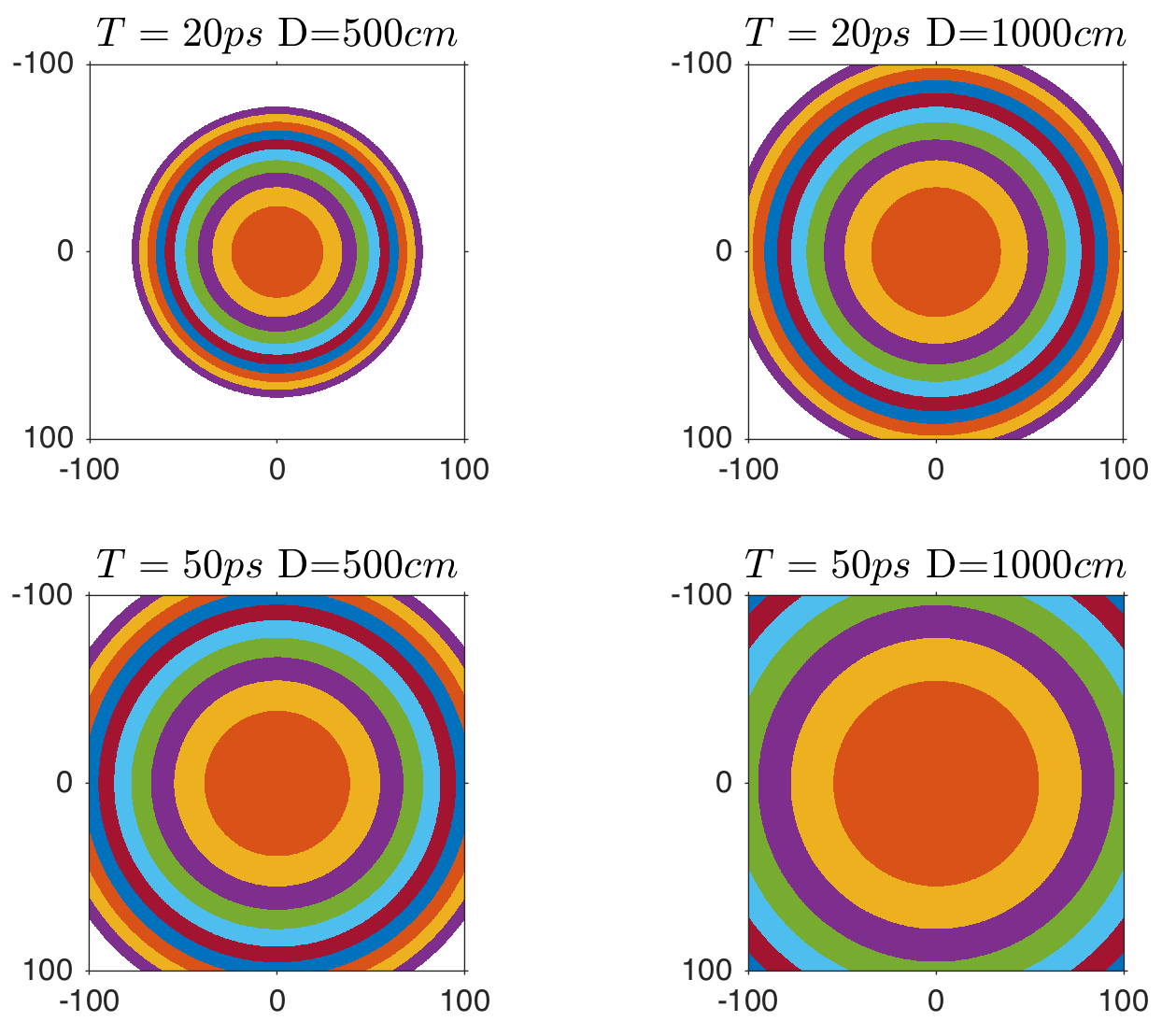}
\caption{Measurement of a planar scene for various sensor time resolutions~$T$ and target distances $D$. The color represents time samples indexes (for the first $10$ samples). As the time resolution worsens or the target is further away, the rings become	 thicker. The images show a subset area of $\SI{100}{\centi\meter} \times \SI{100}{\centi\meter}$.}
\label{fig:ring_structure}
\end{figure}

All the properties presented in the previous subsection extend to the case of a planar scene. Eq. \ref{eq:general_measurement_with_angle} shows that the measurement process integrates over circles centered around the sensor. Due to the finite time resolution, the circles are mapped to rings. The rings are thinner for further points, according to ${{\rho _n} = \sqrt {{c^2}{{\left( {nT + {t_0}} \right)}^2} - {D^2}}} $, where $n$ is the time sample number and $t_0=D/c$ is the time of arrival from the closest point. Fig. \ref{fig:ring_structure} shows the ring structure for a few cases of time resolution and target distance. 

Lastly, Eq. \ref{eq:general_measurement_with_angle} provides the structure of the $\mathbf{H}_i$ matrix, and guidelines for the effects of changing the sensor time resolution and position on the measurement matrix $\mathbf{Q}$. Naturally, better time resolution will reduce the mutual coherence. An alternative to improved time resolution which might be technically challenging is to add more sensors as discussed next.

%%%%%%%%%%%%%%%%%%%%%%%%%%%%%%%%%%%%%%%%%%%%%%%%%%%%%%%%%%%%%%%%%%%%%%%%%%%%%%%%%%%%%%%%%%%%%%%%%%%%%%%%%%%%%%%%%%%%%%%%%%%%%%%%%%%%%%%%%%%%%%%%%%%%%%%%%%%%%%%%%%%%%%%%%%%%%%%%%%%%%%%%%%%%%%
\section{Optimized Time-Resolved Sensor Array}
\label{sec:optimized_array}

Using multiple sensors is a natural extension to the single pixel camera. In the case where the sensors are time sensitive, their positioning affects the measurement matrix $\mathbf{Q}$ and so it can be optimized. Here we derive an algorithm for sensors placement in an array in order to reduce the mutual coherence of $\mathbf{Q}$. To simplify the array structure we constrain the sensors to be placed on a single plane $z=0$ and constrained to an allowed physical area. The algorithm accepts two parameters: the number of sensors $K$ and the allowed physical area $\mathcal{C}$, and provides the ideal positions under these constraints.

Starting with Eq. \ref{eq:general_measurement_with_angle}, the goal is to maximize the difference between $m(x_1,y_1,t)$ and $m(x_2,y_2,t)$. This is achieved by choosing $\mathbf{r}_1=(x_1,y_1)$, $\mathbf{r}_2=(x_2,y_2)$ which are furthest apart (to minimize overlap of the rings as shown in Fig. \ref{fig:ring_structure}). 

More precisely, the goal is to select $i=1..K$ positions $\mathbf{r}_i$ within an area $\mathcal{C}$ such that the distance between the sensors is maximized. This can be achieved by solving:
\begin{equation}
{\left\{ {{\mathbf{r}_i}} \right\}_{i = 1..K}} = \mathop {\arg\max }\limits_{{{\left\{ {{\mathbf{r}_i} \in \mathcal{C}} \right\}}_{i = 1..K}}} \left\{ {\sum\limits_{k = 1}^K {\mathop {\min }\limits_{k \ne k'} \left\| {{\mathbf{r}_k} - {\mathbf{r}_{k'}}} \right\|_2} } \right\}
\label{eq:sensor_optimization}
\end{equation}
Eq.~\ref{eq:sensor_optimization} can be solved by a grid search for a small number of sensors. A more general solution is to relax the problem and follow the equivalent of a Max-Lloyd quantizer~\cite{Peyre2006}. The steps are as follows:
\begin{enumerate}
\item Initialize $K$ random positions in the allowed area $\cal{C}$
\item Repeat until convergence:
\begin{itemize}
\item Calculate the Voronoi diagram of the point set.
\item Move each sensor position to the center of its cell.
\end{itemize}
\end{enumerate}

\begin{figure}[t]
\centering
\includegraphics[width=\linewidth]{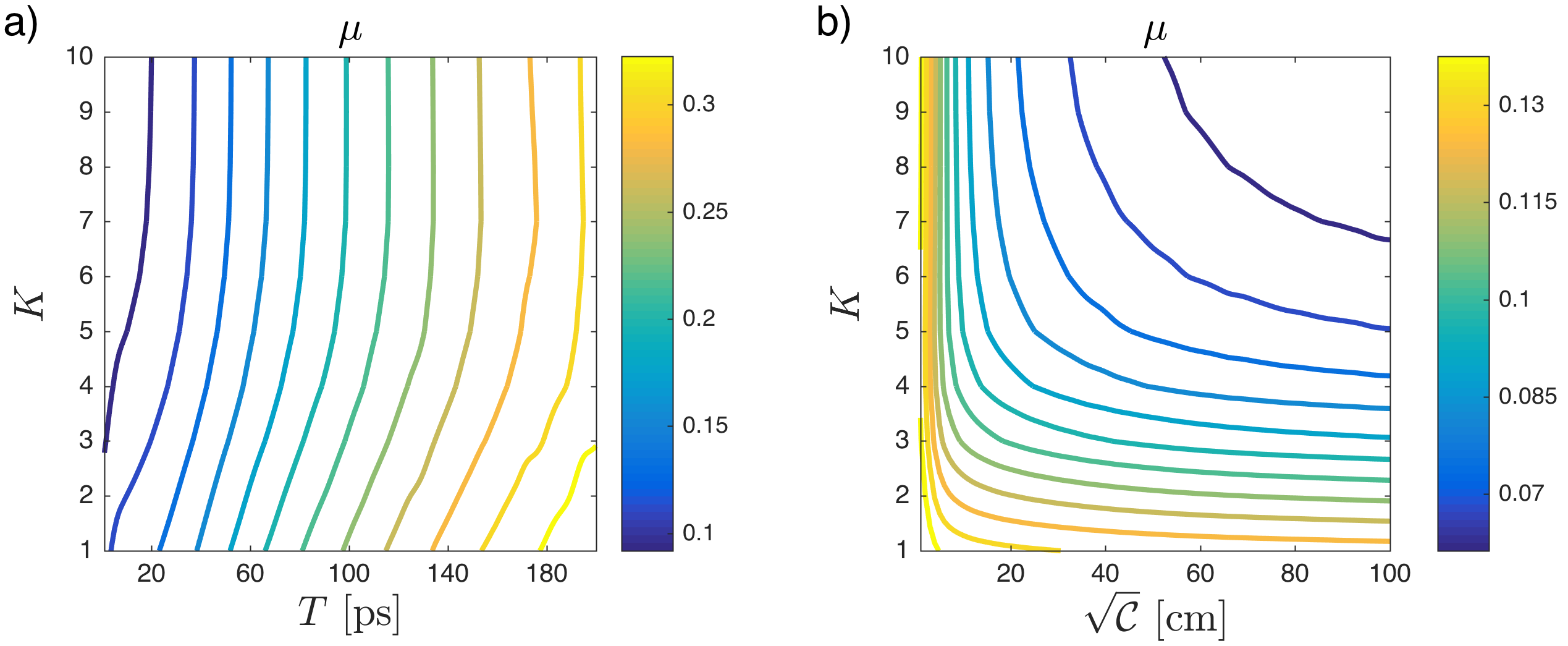}
\caption{Effect of number of sensors $K$, their time resolution $T$, and array size constraint $\mathcal{C}$ on the mutual coherence $\mu$ evaluated with Eq. \ref{eq:cohrence_measure}. The target size is $\SI{5}{\meter} \times \SI{5}{\meter}$, composed of $80 \times 80$ pixels, and at a distance of $D=\SI{10}{\meter}$ from the sensor plane. a) Mutual coherence contours for a varying number of sensors and their time resolution (for fixed array size $\mathcal{C}=\SI{10}{\centi\meter} \times \SI{10}{\centi\meter}$). b) Similar to (a) with varying array size constraint (for fixed time resolution $T=\SI{20}{\pico\second}$).}
\label{fig:coherence_sensor_number}
\end{figure}

This positioning algorithm is evaluated for various system parameters by assessing the effect of the sensor time resolution, number of sensors and array size (square of varying sizes) on the coherence cost objective (Eq. \ref{eq:cohrence_measure}). Fig. \ref{fig:coherence_sensor_number} shows these results. Several key features of the system appear in this analysis: 1) Improving time resolution reduces the number of required sensors non-linearly. 2) It is always beneficial to improve the sensors' time resolution. 3) The sensor area defines a maximum number of useful sensors, beyond which there is no significant decrease in the mutual coherence (increasing the array size linearly reduces the mutual coherence for a fixed number of sensors). 4) It is possible to easily trade off between different aspects of the system's hardware by traveling on the contours. For example, a decrease in the sensor time resolution can be balanced by adding more sensors. This can be useful for realizing an imaging system as sensors with lower time resolution are less expensive and easier to manufacture. The next section provides an alternative to improving hardware by adding structured illumination.

%%%%%%%%%%%%%%%%%%%%%%%%%%%%%%%%%%%%%%%%%%%%%%%%%%%%%%%%%%%%%%%%%%%%%%%%%%%%%%%%%%%%%%%%%%%%%%%%%%%%%%%%%%%%%%%%%%%%%%%%%%%%%%%%%%%%%%%%%%%%%%%%%%%%%%%%%%%%%%%%%%%%%%%%%%%%%%%%%%%%%%%%%%%%%%
\section{Optimized Active Illumination for Time-Resolved Sensor Array}
\label{sec:optimized_illum}

We now make the leap to compressive sensing. Previous sections discussed single sensor considerations and sensors placement in an array. This section covers ideal active illumination patterns. We assume the illumination wavefront is amplitude-modulated; this can be physically achieved by an SLM or liquid crystal display (LCD). 

When considering different illumination patterns, Hadamard patterns and random patterns sampled from Bernoulli distribution are normally chosen. Instead, we suggest patterns that directly aim to minimize the mutual coherence of the measurement matrix. The mathematical patterns may have negative values which can be represented by taking a measurement with an ``all on'' pattern and subtracting it from the other measurements (due to the linearity of the system)~\cite{Bahmani2015}.

In order to optimize the illumination patterns, we follow the proposal in~\cite{Duarte-Carvajalino2009} and learn the sensing matrix (illumination patterns) in order to directly minimize the coherence of the measurement matrix. For example, this concept has been reduced to practice in~\cite{Marwah2013}. 

The crux of the idea is that given a number of allowed illumination patterns $M$, we choose the set of illumination patterns that minimizes the mutual coherence of the matrix~$\mathbf{Q}$. Since the illumination matrix $\mathbf{G}_j$ is performing pixel-wise modulation of the target, it is a diagonal matrix with the pattern values on the diagonal ${\mathbf{G}_j} = diag\{ {\mathbf{g}_j}\} $, where $\mathbf{g}_j$ is a vector containing the $j$-th pattern values. Taking a closer look at Eq.~\ref{eq:mesurement_operator}, we stack all the sensor matrices $\mathbf{H}_i$ into $ \mathbf{\overline H}$ such that:
\begin{equation}
\mathbf{Q} = \left[ {\begin{array}{*{20}{c}}
 \vdots \\
{{\mathbf{H}_i}{\mathbf{G}_j}}\\
 \vdots 
\end{array}} \right] = \left[ {\begin{array}{*{20}{c}}
{\mathbf{\overline H} {\mathbf{G}_1}}\\
 \vdots \\
{\mathbf{\overline H} {\mathbf{G}_M}}
\end{array}} \right] = \left[ {\begin{array}{*{20}{c}}
{\mathbf{\overline H}  \times diag\{ {\mathbf{g}_1}\} }\\
 \vdots \\
{\mathbf{\overline H} \times diag\{ {\mathbf{g}_M}\} }
\end{array}} \right]
\label{eq:sensor_operator}
\end{equation}
Based on Eq. \ref{eq:cohrence_measure}, the ideal patterns are the solution to:
\begin{equation}
{\{ {\mathbf{g}_j}\} _{j = 1..M}} = \mathop {\arg\min }\limits_{{{\left\{ {{\mathbf{g}_j}\in {{[ - 1,1]}^{{L}}} } \right\}}_{j = 1..M}}} \left\{ {\left\| {{\mathbf{I}_{{L}}} - {{\mathbf{\tilde Q}}^T}\mathbf{\tilde Q}} \right\|_F^2} \right\}
\label{eq:illum_optimization_problem}
\end{equation}
This can be solved with standard constrained optimization solvers. Appendix \ref{sec:app_illum_pattern_optimization} provides the derivation for the cost function and its gradient.

\begin{figure}[t]
\centering
\includegraphics[width=\linewidth]{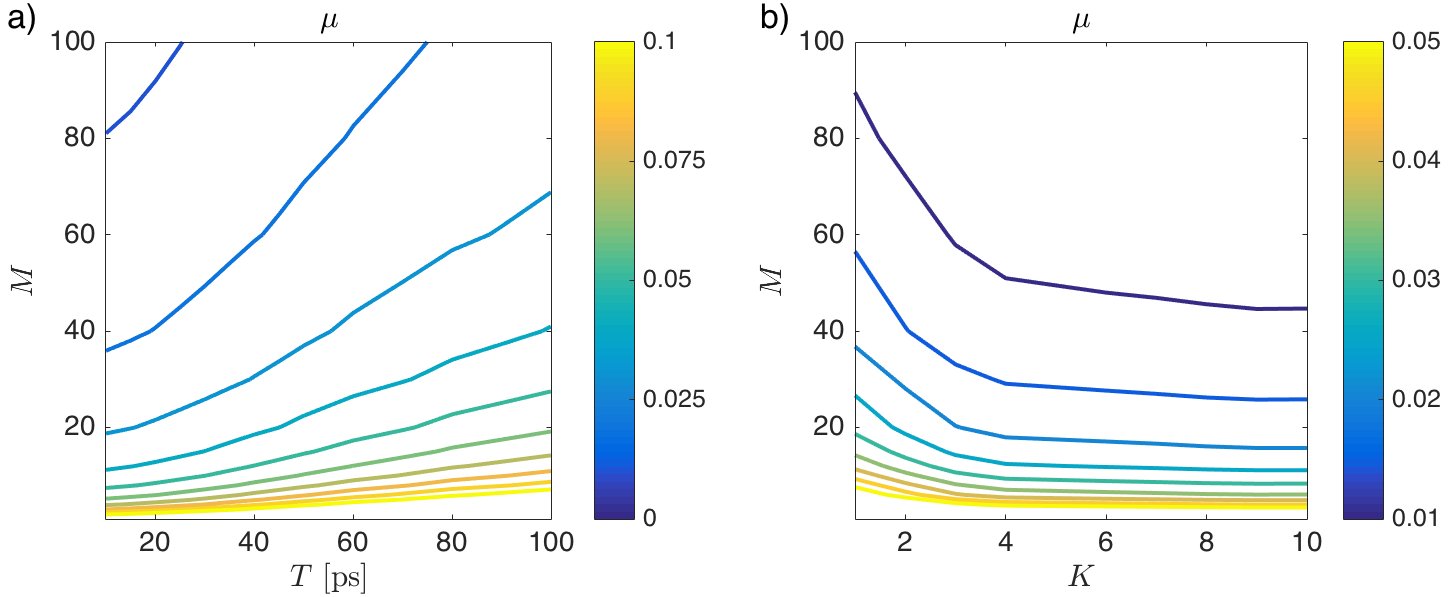}
\caption{Effect of number of illumination patterns $M$, sensor time resolution $T$ and number of sensors $K$ on the mutual coherence $\mu$ evaluated with Eq.~\ref{eq:cohrence_measure}. The target size is $\SI{5}{\meter} \times \SI{5}{\meter}$, composed of $80 \times 80$ pixels, and at a distance of $D=\SI{10}{\meter}$ from the sensor plane. The sensor area $\mathcal{C}$ is a square of size ${\SI{10}{\centi\meter} \times \SI{10}{\centi\meter}}$. a)~Mutual coherence contours for a varying number of illumination patterns and sensors' time resolution (for a fixed number of sensors $K=1$). b)~Similar to (a) with a varying number of sensors (for fixed time resolution $T=\SI{20}{\pico\second}$).}
\label{fig:coherence_lum_patterns_sensors}
\end{figure}

Fig. \ref{fig:coherence_lum_patterns_sensors} shows the change in mutual coherence in simulations while varying the number of allowed illumination patterns, the sensor time resolution, and the number of sensors. As predicted by CS theory, increasing the number of patterns has a strong effect on the mutual coherence. This strong effect allows one to easily relax the demands on the hardware requirements when needed. However, as more patterns are allowed, there are increasingly more dependencies on the sensors' parameters. This demonstrates the synergy between compressive sensing and time-resolved sensing. In this case traveling on mutual coherence contours allows one to trade-off system complexity (cost, size, power) with acquisition time (increased when more patterns are required).

\begin{figure}[t]
\centering
\includegraphics[width=\linewidth]{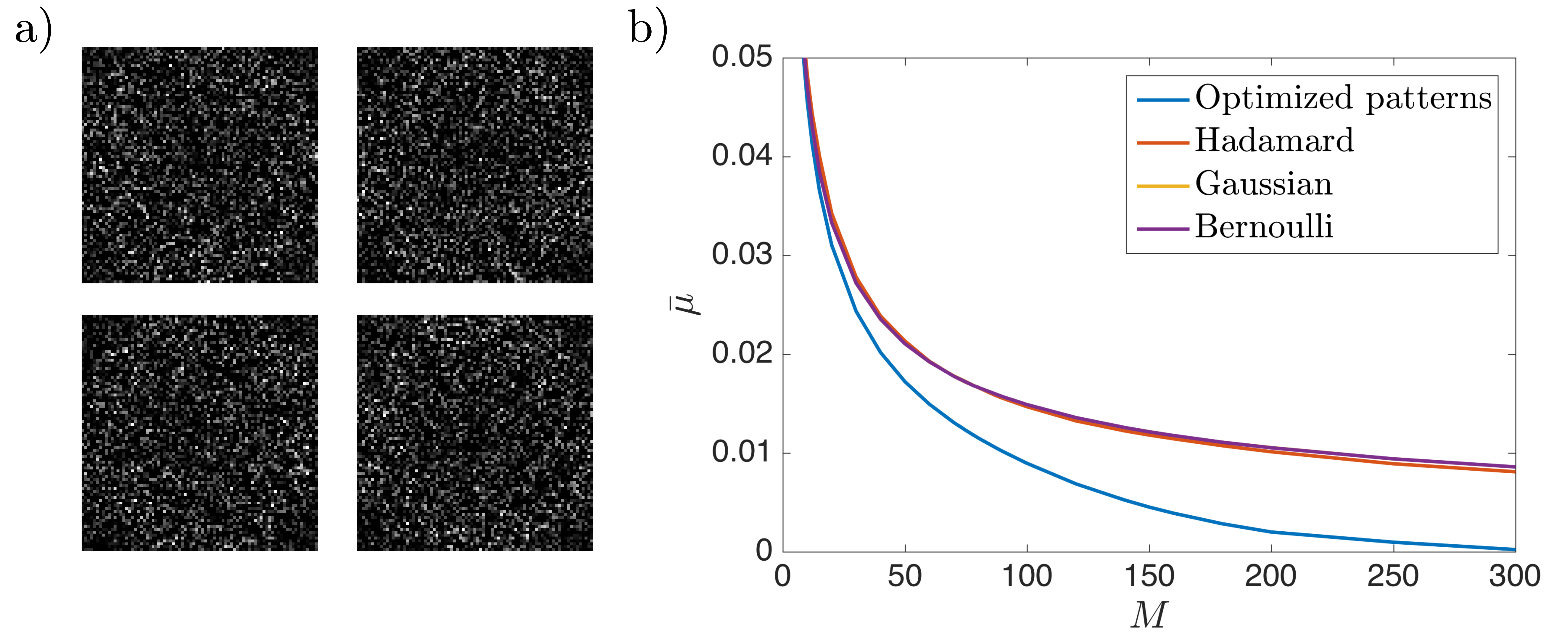}
\caption{The value of optimized active illumination patterns. The patterns are optimized for a $\SI{5}{\meter} \times \SI{5}{\meter}$ target composed of $80 \times 80$ pixels at a distance of $D=\SI{10}{\meter}$ from the sensor plane. The measurement is simulated with $K=1$ sensors and $T=\SI{20}{\pico\second}$. a) Examples of several patterns computed for ${M=50}$. b) Comparison of different active illumination methods and their effect on the mutual coherence for varying $M$. The optimized patterns outperform Hadamard and random patterns sampled from Gaussian and Bernoulli (in $\left\{ { - 1,1} \right\}$) distributions. }
\label{fig:comparison_ilum_pats}
\end{figure}

Fig.~\ref{fig:comparison_ilum_pats}a shows several examples of the patterns computed by solving Eq.~\ref{eq:illum_optimization_problem}. Fig.~\ref{fig:comparison_ilum_pats}b demonstrates the value of the optimized patterns compared to Hadamard and random patterns sampled from Gaussian and Bernoulli (in $\left\{ { - 1,1} \right\}$) distributions. For very few illumination patterns (below ten) all patterns are comparable. When allowing more illumination patterns, the optimized patterns are performing better by reducing the mutual coherence faster compared to the other approaches. As predicted, the performances of Hadamard, Gaussian and Bernoulli patterns are nearly identical. 

\begin{figure}[t]
\centering
\includegraphics[width=\linewidth]{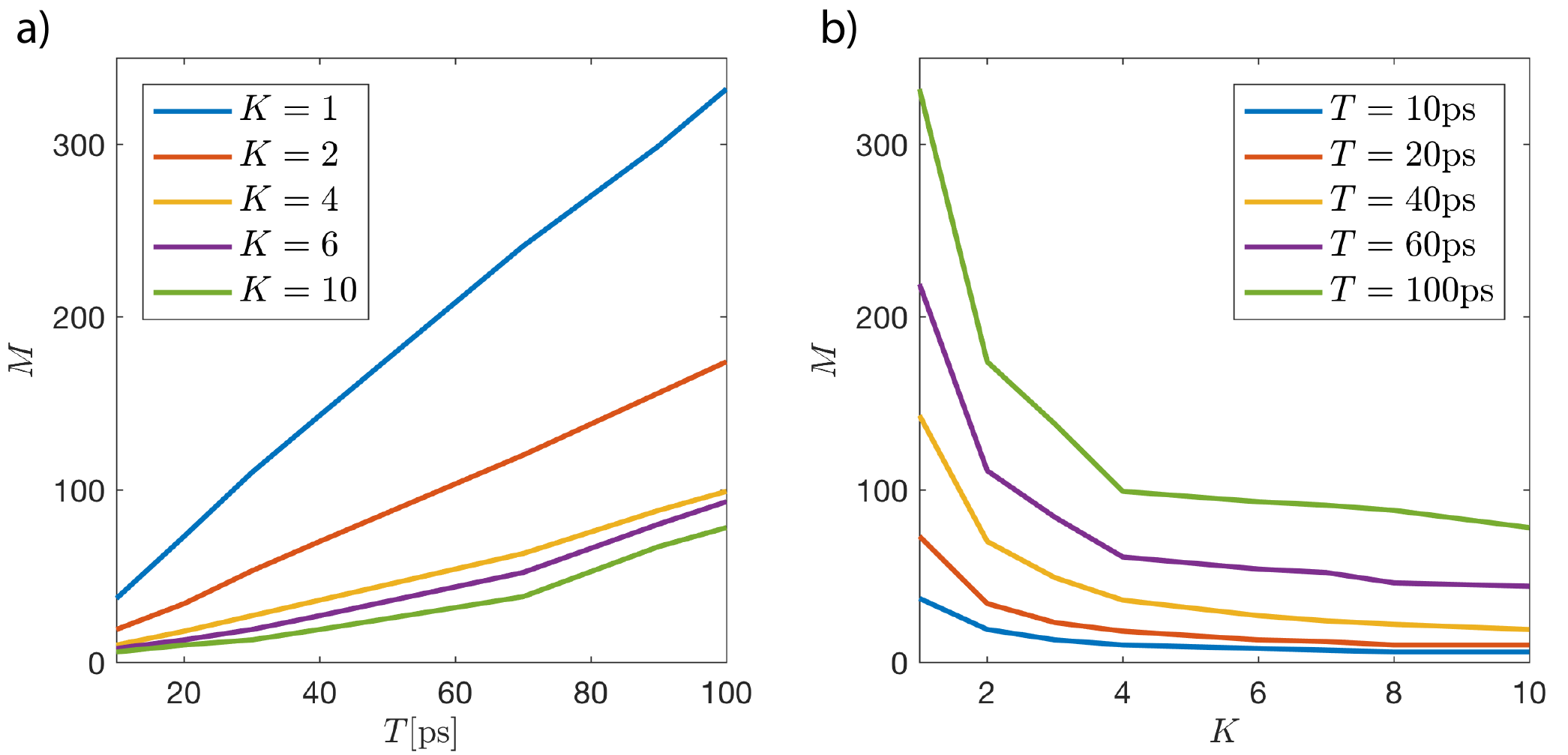}
\caption{Effect of system parameters on reconstruction quality. Various design points (different number of sensors $K$ and time resolution $T$) are simulated. The number of optimized illumination patterns $M$ is set as the minimal number of patterns required to achieve reconstruction quality with SSIM~$\geq 0.95$ and PSNR~$\geq \SI{40}{\deci\bel}$. The target used is the cameraman image (see Fig.~\ref{fig:numerical_reconstruction} right). a) Demonstrate the trends of various number of detectors $K$ as a function of the time resolution $T$. b) Shows the trends of different detector time resolution as a function of the number of detectors. }
\label{fig:ssim_for_num_illum_pats}
\end{figure}

%%%%%%%%%%%%%%%%%%%%%%%%%%%%%%%%%%%%%%%%%%%%%%%%%%%%%%%%%%%%%%%%%%%%%%%%%%%%%%%%%%%%%%%%%%%%%%%%%%%%%%%%%%%%%%%%%%%%%%%%%%%%%%%%%%%%%%%%%%%%%%%%%%%%%%%%%%%%%%%%%%%%%%%%%%%%%%%%%%%%%%%%%%%%%%
\section{Numerical Result}

\begin{figure*}[t]
\centering
\includegraphics[width=\linewidth]{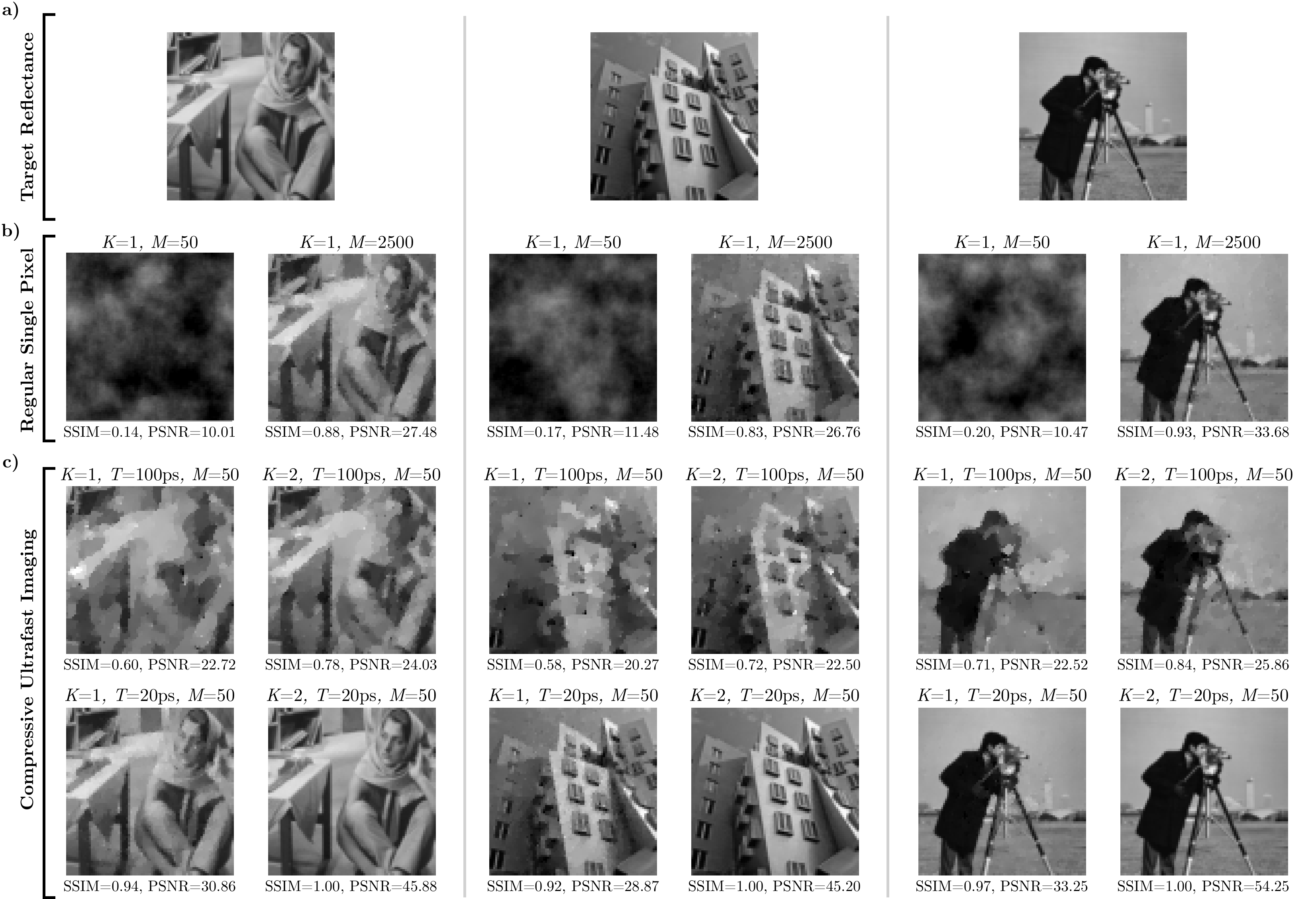}
\caption{
Imaging with compressive ultrafast sensing for different targets. 
a) The target image. 
b) Result with regular single pixel camera with $M=50$ and $M=2500$. 
c) Results with compressive ultrafast sensing with $M=50$ for four design points with time resolution of $T=\SI{100}{\pico\second}$ and $T=\SI{20}{\pico\second}$, and  $K=1$ and $K=2$. All reconstructions are evaluated with SSIM and PSNR. The results demonstrate the strong dependency on time resolution. Result for $K=2$ and $T=\SI{20}{\pico\second}$ shows perfect reconstruction on all targets based on SSIM. All measurements were added with white Gaussian noise such that the measurement SNR is $\SI{60}{\deci\bel}$.
}
\label{fig:numerical_reconstruction}
\end{figure*}

This section demonstrates target reconstruction using the above analysis. The target dimensions are $\SI{5}{\meter} \times \SI{5}{\meter}$ with ${80 \times 80}$ pixels ($L=6400$) and it is placed $\SI{10}{\meter}$ away from the detector plane. The detector array is limited to a square area of $\mathcal{C}=\SI{10}{\centi\meter} \times \SI{10}{\centi\meter}$. The detector placement method used is described in section~\ref{sec:optimized_array} and the illumination patterns are computed using the algorithm suggested in section~\ref{sec:optimized_illum}. The measurement operator is simulated as described in section~\ref{sec:time_resolved_light_transport} to produce the total measurement vector. White Gaussian noise is added to the total measurement vector to produce a measurement SNR of $\SI{60}{\deci\bel}$. 
The targets simulated here are natural scenes (sparse in gradient domain). To invert Eq. \ref{eq:mesurement_operator} we use TVAL3~\cite{li2009user} (with TVL2 and a regularization parameter of $2^{13}$ for all targets). The reconstruction quality is evaluated with both Peak Signal to Noise Ratio (PSNR --- higher is better, performs pointwise comparison) and Structural Similarity index (SSIM --- ranges in $[0,1]$, higher is better, takes into account the spatial structure of the image~\cite{Wang2004}).

So far, the discussion focused on reducing the mutual coherence of the measurement matrix $\mathbf{Q}$. Fig.~\ref{fig:ssim_for_num_illum_pats} demonstrates the effect on the full reconstruction process. The target used is the cameraman image (Fig.~\ref{fig:numerical_reconstruction} right). The goal is to find the minimal number of illumination patterns in order to produce a reconstruction quality defined by SSIM~$\geq 0.95$ and PSNR~$\geq \SI{40}{\deci\bel}$. This is repeated for various number of detectors with different time resolutions. The trends demonstrate a linear relationship between the number of illumination patterns and the detector time resolution needed for a specified reconstruction quality. Another notable effect is the significant gain in the transition from one to two detectors followed by a diminishing gain for additional detectors. This gain decreases as the detector time resolution improves. These trends can be useful to trade off design constraints. For example, for the specified reconstruction quality the user can choose one detector with a time resolution of $\SI{20}{\pico\second}$ and 80 patterns. The same acquisition time can be maintained with two simpler detectors of $\SI{40}{\pico\second}$. Alternatively, two detectors with $\SI{20}{\pico\second}$ require only 40 patterns (shorter acquisition time) for equal reconstruction quality.

Finally, we compare the suggested design framework to a traditional (non-time aware) single pixel camera. This is simulated with an $\mathbf{H}$ matrix with just one row with ones. The illumination patterns are sampled from a Bernoulli random distribution in $\left\{ { - 1,1} \right\}$ in a similar way to the original single pixel camera experiments~\cite{Duarte2008a}. Fig.~\ref{fig:numerical_reconstruction} shows the results for three different targets. Reconstructions with a traditional single pixel camera are shown in Fig.~\ref{fig:numerical_reconstruction}b for ${M=50}$ and ${M=2500}$ patterns. Four different design points of compressive ultrafast imaging are demonstrated in Fig.~\ref{fig:numerical_reconstruction}c: ${\left\{ {K = 1,T = \SI{100}{\pico\second}} \right\}}$, ${\left\{ {K = 2,T = \SI{100}{\pico\second}} \right\}}$, ${\left\{ {K = 1,T = \SI{20}{\pico\second}} \right\}}$, and ${\left\{ {K = 2,T = \SI{20}{\pico\second}} \right\}}$, all with ${M=50}$ patterns (such that the acquisition time is equal). Several results are worth noting: 
\begin{itemize}
\item Reconstruction with ${K=2}$, ${T=\SI{20}{\pico\second}}$, and ${M=50}$ achieves perfect quality based on SSIM for all targets.
\item Reconstruction with ${K=1}$, ${T=\SI{20}{\pico\second}}$, and ${M=50}$ outperforms the traditional single pixel camera approach with $50 \times$ fewer illumination patterns and demonstrates the potential gain of this approach.
\item A traditional single pixel reconstruction with ${M=50}$ patterns (same acquisition time as the compressive ultrafast imaging design points discussed) fails to recover the scene information.
\item There is a significant gain in performance when improving the sensor time resolution.
\end{itemize}

%%%%%%%%%%%%%%%%%%%%%%%%%%%%%%%%%%%%%%%%%%%%%%%%%%%%%%%%%%%%%%%%%%%%%%%%%%%%%%%%%%%%%%%%%%%%%%%%%%%%%%%%%%%%%%%%%%%%%%%%%%%%%%%%%%%%%%%%%%%%%%%%%%%%%%%%%%%%%%%%%%%%%%%%%%%%%%%%%%%%%%%%%%%%%%
\section{Discussion}
Section~\ref{sec:optimized_illum} analyzed only wavefront amplitude modulation. There are many other ways to use coded active illumination in order to minimize the measurement coherence. For example, we assumed the wavefront is just a pulse in time, but we can perform coding in time domain as well. This will cause different pixels on the target image to be illuminated at different times. Physical implementation of such delays is possible with, for example, tilted illumination and fiber bundles (notice that while phase SLM induces varying time delays on the wavefront, these time scales are shorter than current time-resolved sensors resolution). Analysis of such implementation requires detailed care with the interplay between the $\mathbf{H}$ and $\mathbf{G}$ matrices (since $\mathbf{G}$ becomes time-dependent); we leave this analysis to a future study.

The forward model (Eq.~\ref{eq:general_measurement}) assumes the wave nature of light is negligible. This assumption is valid if 1) Diffraction is negligible: the scene spatial features are significantly greater compared to the illumination wavelength (order of~$\SI{}{\micro\meter}$). 2)~Interference is negligible: the coherence length of the illumination source is significantly smaller compared to the geometrical features. For pulsed lasers the coherence length is inversely proportional to the pulse bandwidth; this usually results in sub-$\SI{}{\centi\meter}$ coherence lengths.

The suggested approach provides a framework for lensless imaging with compressive ultrafast sensing. This framework provides the user with design tools for situations in which lensless imaging is essential. It allows the user to effectively balance available resources --- an important tool since the hardware requirements can be substantial (pulsed source with structured illumination and time-resolved sensors). We note that time-resolved sensors are becoming more accessible with the recent advances in CMOS based SPAD devices (e.g.~\cite{Richardson2009}). Another limitation of our approach is the requirement of known geometry. Interestingly, the approach suggested in~\cite{Kirmani2011,colacco2012compressive} requires similar hardware to recover scene geometry without reflectance, hence it might be possible to fuse the two approaches in the future.

%%%%%%%%%%%%%%%%%%%%%%%%%%%%%%%%%%%%%%%%%%%%%%%%%%%%%%%%%%%%%%%%%%%%%%%%%%%%%%%%%%%%%%%%%%%%%%%%%%%%%%%%%%%%%%%%%%%%%%%%%%%%%%%%%%%%%%%%%%%%%%%%%%%%%%%%%%%%%%%%%%%%%%%%%%%%%%%%%%%%%%%%%%%%%%
\section{Conclusion}
We demonstrated a novel compressive imaging architecture for using ultrafast sensors with active illumination for lensless imaging. We discussed analysis tools for hardware design, as well as algorithms for ideal sensor placement and illumination patterns which directly target the RIP for robust inversion with compressive deconvolution. The presented approach allows lensless imaging with single pixel and dramatically better acquisition times compared to previous results. This enables novel lensless single pixel imaging in challenging environments. The approach and analysis presented here open new avenues for other areas with potential tight coupling between novel sensors and compressive sensing algorithms.

% or


\appendices

\section{Illumination Pattern Optimization Algorithm}
\label{sec:app_illum_pattern_optimization}

Here we provide a derivation for calculating the cost function in Eq. \ref{eq:illum_optimization_problem} and its gradient. Starting with the cost function to minimize:
\begin{equation}
\gamma  = \left\| {{\mathbf{I}_{{L}}} - {{ \mathbf{\tilde Q}}^T} \mathbf{\tilde Q}} \right\|_F^2
\label{eq:a:cost1}
\end{equation} 
Define $\pmb{\Lambda}$ such that its $j$-th row is $\mathbf{g}_j^T$ ($\pmb{\Lambda}$ is an $M \times L$ matrix). Our goal is to find $\pmb{\Lambda}$ which minimizes $\gamma$. 

We start by writing: $\mathbf{O} = \mathbf{\tilde Q}^T \mathbf{\tilde Q}$, and so:
\begin{equation}
\begin{split}
\gamma  & = {\left\| {{\mathbf{I}_{{L}}} - \mathbf{O}} \right\|_F^2} = Tr\left\{ {\left( {{\mathbf{I}_{{L}}} - \mathbf{O}} \right){{\left( {{\mathbf{I}_{{L}}} - \mathbf{O}} \right)}^T}} \right\}  \\
& = Tr\left\{ \mathbf{I}_L \right\} - 2Tr\left\{ \mathbf{O} \right\} + Tr\left\{ {\mathbf{O}{\mathbf{O}^T}} \right\} =  {\left\| \mathbf{O} \right\|_F^2} - {L}
\end{split}
\label{eq:a:norm_def}
\end{equation}
since $Tr\left\{ \mathbf{O} \right\}=L$. Next, we define $\mathbf{\tilde Q}=\mathbf{QP}$ where $\mathbf{P}$ is a diagonal matrix with the inverse of the columns norm: 
\begin{equation}
\mathbf{P}_{n,n}=\frac{1}{{\sqrt {\sum\limits_{a = 1}^{MKN} {{\mathbf{Q}_{a,n}}^2} } }} = \frac{1}{{\sqrt {{\left( {\sum\limits_{j = 1}^M {{{\mathbf{\Lambda }}_{j,n}}^2} } \right)\left( {\sum\limits_{i = 1}^{KN} {{{\mathbf{\bar H}}_{i,n}}^2} } \right)}} }}
\label{eq:a:column_weight}
\end{equation}
This allows us to write:
\begin{equation}
\mathbf{O} = \mathbf{\tilde Q}^T \mathbf{\tilde Q} = \mathbf{P}^T \mathbf{Q}^T \mathbf{QP} = \mathbf{P} \sum\limits_{j = 1}^M {\left( {{\mathbf{D}_{{\mathbf{\Lambda} _j}}}\mathbf{\bar H}^T \mathbf{\bar H}{\mathbf{D}_{{\mathbf{\Lambda} _j}}}} \right)}  \mathbf{P}
\label{eq:a:break_gramm}
\end{equation}
where $\mathbf{D}_{{\mathbf{\Lambda}}_j}$ is a diagonal matrix with the $j$-th row of $\mathbf{\Lambda}$ on the diagonal. Next we note that: 
\begin{equation}
{\left[ {\mathbf{\bar H{D_{{\Lambda} _j}}}} \right]_{n,m}} = \sum\limits_{a = 1}^{L} {{\mathbf{\bar H}_{n,a}}{{\left[ {{\mathbf{D_{{\Lambda} _j}}}} \right]}_{a,m}}}  = {\mathbf{\bar H}_{n,m}}{\mathbf{\Lambda} _{j,m}}
\label{eq:a:matrix_break}
\end{equation}
and:
\begin{equation}
{\mathbf{O}_{n,m}} = {\mathbf{P}_{n,n}}\sum\limits_{j = 1}^M {\left( {{\mathbf{\Lambda} _{j,n}}{{\left[ \mathbf{{{\bar H}^T \bar H}} \right]}_{n,m}}{\mathbf{\Lambda} _{j,m}}} \right)} {\mathbf{P}_{m,m}}
\label{eq:a:O_mn}
\end{equation}
which can be simplified to:
 \begin{equation}
{\mathbf{O}_{n,m}} = {\mathbf{W}_{n,m}}\frac{{\sum\limits_{j = 1}^M {{\mathbf{\Lambda} _{j,n}}{\mathbf{\Lambda} _{j,m}}} }}{{\sqrt {\left( {\sum\limits_{j = 1}^M {{\mathbf{\Lambda} _{j,n}}^2} } \right)\left( {\sum\limits_{j = 1}^M {{\mathbf{\Lambda} _{j,m}}^2} } \right)} }}
 \label{eq:a:O_mn_final}
 \end{equation}
with
\begin{equation}
{\mathbf{W}_{n,m}} = \frac{{{{\left( {\mathbf{\bar H}^T \mathbf{\bar H}} \right)}_{n,m}}}}{{\sqrt {\left( {\sum\limits_{i = 1}^K {{\mathbf{H}_{i,n}}^2} } \right)\left( {\sum\limits_{i = 1}^K {{\mathbf{H}_{i,m}}^2} } \right)} }}
\label{eq:a:W_m,n}
\end{equation}
Lastly, we define: 
\begin{equation}
\mathbf{\Phi}=\mathbf{\Lambda}^T \mathbf{\Lambda}
\label{eq:a:phi_def}
\end{equation} 
such that:
\begin{equation}
{\mathbf{O}_{n,m}} = {\mathbf{W}_{n,m}}\frac{{{\mathbf{\Phi} _{n,m}}}}{{\sqrt {{\mathbf{\Phi} _{n,n}}{\mathbf{\Phi} _{m,m}}} }}
\end{equation}
which allows us to write:
\begin{equation}
\mathbf{O}={\left( {\sqrt {{{\mathbf{S}}_{\mathbf{\Phi }}}} } \right)^{ - 1}}  \left( {\mathbf{W} \odot \mathbf{\Phi} } \right)  {\left( {\sqrt {{{\mathbf{S}}_{\mathbf{\Phi }}}} } \right)^{ - 1}}
\label{eq:a:final_O}
\end{equation}
where $\mathbf{S_\Phi}$ is a diagonal matrix with the diagonal entries of $\mathbf{\Phi}$, and $\odot$ denotes element wise multiplication. Finally:
\begin{equation}
\gamma={\left\| {\left( {\sqrt {{{\mathbf{S}}_{\mathbf{\Phi }}}} } \right)^{ - 1}}  \left( {\mathbf{W} \odot \mathbf{\Phi} } \right)  {\left( {\sqrt {{{\mathbf{S}}_{\mathbf{\Phi }}}} } \right)^{ - 1}} \right\|_F^2} - {L}
\label{eq:a:final_eq_cost}
\end{equation}
We note that $\mathbf{W}$ (Eq. \ref{eq:a:W_m,n}) is a constant matrix for the illumination pattern optimization and can be calculated a priori. Eq.~\ref{eq:a:phi_def} and~\ref{eq:a:final_eq_cost} provide the final expression for $\gamma(\mathbf{\Lambda})$.

We now develop an expression for the gradient of the cost function. Considering the chain rule for matrices~\cite{petersen2008matrix}:
\begin{equation}
{\left[ {\frac{{d\gamma }}{{d\mathbf{\Lambda} }}} \right]_{n,m}} = \sum\limits_{a = 1}^{{L}} {\sum\limits_{b = 1}^{{L}} {\frac{{d\gamma (\mathbf{\Phi} )}}{{d{\mathbf{\Phi} _{a,b}}}}\frac{{d{\mathbf{\Phi} _{a,b}}}}{{d{\mathbf{\Lambda} _{n,m}}}}} } 
\label{eq:a:chain_rule}
\end{equation}
Starting with the first term, if ${a \ne b}$:
\begin{equation}
\frac{{d\gamma }}{{d{\mathbf{\Phi} _{a,b}}}} = {{\mathbf{W}_{a,b}}^2\frac{{2{\mathbf{\Phi} _{a,b}}}}{{{\mathbf{\Phi} _{a,a}}{\mathbf{\Phi} _{b,b}}}}}
\label{eq:a:grad_1}
\end{equation}
and, if $a=b$:
\begin{equation}
\frac{{d\gamma }}{{d{\mathbf{\Phi} _{a,a}}}} = { - \sum\limits_{c = 1}^n {\frac{{{\mathbf{W}_{a,c}}^2{\mathbf{\Phi} _{a,c}}^2 + {\mathbf{W}_{c,a}}^2{\mathbf{\Phi} _{c,a}}^2}}{{{\mathbf{\Phi} _{a,a}}^2{\mathbf{\Phi} _{c,c}}}}{\delta _{c \ne a}}} } 
\label{eq:a:grad_2}
\end{equation}
where $\delta$ is Kronecker delta. The second term in Eq. \ref{eq:a:chain_rule} is given by~\cite{petersen2008matrix}:
\begin{equation}
\frac{{d{\mathbf{\Phi} _{a,b}}}}{{d{\mathbf{\Lambda} _{n,m}}}} = {\delta _{mb}}{\mathbf{\Lambda} _{n,a}} + {\delta _{ma}}{\mathbf{\Lambda} _{n,b}}
\label{eq:a:grad_2a}
\end{equation}
Combining Eqs. \ref{eq:a:chain_rule} through \ref{eq:a:grad_2a} we get:
\begin{equation}
\begin{split}
{\left[ {\frac{{d\gamma }}{{d{\mathbf{\Lambda }}}}} \right]_{n,m}} & = \sum\limits_{a = 1}^{{L}} {\sum\limits_{b = 1}^{{L}} {{E^1}{\delta _{ab}}\left( {{\delta _{mb}}{{\mathbf{\Lambda }}_{n,a}} + {\delta _{ma}}{{\mathbf{\Lambda }}_{n,b}}} \right)} } \\
& + \sum\limits_{a = 1}^{{L}} {\sum\limits_{b = 1}^{{L}} {{E^2}{\delta _{a \ne b}}\left( {{\delta _{mb}}{{\mathbf{\Lambda }}_{n,a}} + {\delta _{ma}}{{\mathbf{\Lambda }}_{n,b}}} \right)} } 
\end{split}
\label{eq:a:grad_3}
\end{equation}
where $E^1$ and $E^2$ are the terms in Eqs. \ref{eq:a:grad_1} and \ref{eq:a:grad_2} respectively. After some algebra we get:
\begin{equation}
\begin{aligned}
{\left[ {\frac{{d\gamma }}{{d{\mathbf{\Lambda }}}}} \right]_{nm}   } & = \frac{2}{{{\mathbf{\Phi}_{mm}}}} \\
& \sum\limits_{\substack{a = 1\\a \ne m}}^{{L}} [ \frac{{{\mathbf{\Lambda}_{na}}}}{{{\mathbf{\Phi}_{aa}}}}\left( {{\mathbf{W}_{am}}^2{\mathbf{\Phi}_{am}} + {\mathbf{W}_{ma}}^2{\mathbf{\Phi}_{ma}}} \right) - \\
& \frac{{{\mathbf{\Lambda}_{nm}}}}{{{\mathbf{\Phi}_{aa}}{\mathbf{\Phi}_{mm}}}}\left( {{\mathbf{W}_{am}}^2{\mathbf{\Phi}_{am}}^2 + {\mathbf{W}_{ma}}^2{\mathbf{\Phi}_{ma}}^2} \right) ]
\end{aligned}
\label{eq:a:grad_4}
\end{equation}
Lastly, we define:
\begin{equation}
\begin{split}
{{\mathbf{\alpha }}_1} & = {\mathbf{W}} \odot {\mathbf{W}} \odot {\mathbf{\Phi }} \odot {\mathbf{I}^ - } \\
{\mathbf{\alpha} _2} & = {\mathbf{W}} \odot {\mathbf{W}} \odot {\mathbf{\Phi }} \odot {\mathbf{\Phi }} \odot {\mathbf{I}^ - } \\
\end{split}
\label{eq:a:final_grad_const}
\end{equation}
where, $\mathbf{I}^ - = 1 - \mathbf{I}_{L}$ (matrix with all ones except for zeros on the diagonal), which allows us to write the final gradient as:
\begin{equation}
\begin{split}
\mathbf{c}_1 & = {\Lambda {{\bf{S}}_{\bf{\Phi }}}^{ - 1}\left( {{\mathbf{\alpha} _1} + {\mathbf{\alpha} _1}^T} \right)} \\
\mathbf{c}_2 & = {\left( {\mathbf{\Lambda}  \odot \left( {{\mathbf{1}_{M \times L}}{{\bf{S}}_{\bf{\Phi }}}^{ - 1}\left( {{\mathbf{\alpha} _2} + {\mathbf{\alpha} _2}^T} \right){{\bf{S}}_{\bf{\Phi }}}^{ - 1}} \right)} \right)} \\
\frac{{d\gamma }}{{d{\bf{\Lambda }}}} & = 2\left( \mathbf{c}_1-\mathbf{c}_2 \right){{\bf{S}}_{\bf{\Phi }}}^{ - 1}
\end{split}
\end{equation}
where $\mathbf{1}_{M \times L}$ is an $M \times L$ matrix with all ones.


\ifCLASSOPTIONcaptionsoff
  \newpage
\fi



\bibliographystyle{IEEEtran}
\bibliography{IEEEabrv,LenslessImaging}
%\begin{thebibliography}{1}
%\end{thebibliography}


%

%\newpage


%\vfill


\end{document}